\documentclass[3p,times,procedia,final]{elsarticle}
\pdfoutput = 1
\usepackage[procedia]{ecrc}
\usepackage[bookmarks=false]{hyperref}
    \hypersetup{colorlinks,
      linkcolor=blue,
      citecolor=blue,
      urlcolor=blue}

\volume{192}

\firstpage{961}

\journalname{Procedia Computer Science}

\runauth{E. Mengin and F. Rossi}

\jid{procs}

\usepackage{amssymb}
\usepackage[figuresright]{rotating}

\usepackage{amsmath}
\usepackage{amsfonts}
\usepackage{amssymb}
\usepackage{amsthm}
\usepackage{nicefrac}

\usepackage{graphicx}
\usepackage{subcaption}
\usepackage{multirow}
\usepackage{booktabs}

\usepackage{todonotes}
\usepackage[english]{babel}
\usepackage{microtype}

\DeclareMathOperator*{\argmax}{argmax}

\begin{document}
\begin{frontmatter}

\dochead{25th International Conference on Knowledge-Based and Intelligent Information \& Engineering Systems}%

\title{Improved Algorithm for the Network Alignment Problem with Application to Binary Diffing}

\author[a, b]{Elie Mengin \corref{cor1}} 
\author[c]{Fabrice Rossi}

\address[a]{SAMM, EA 4543, Universit\'e Paris 1 Panth\'eon-Sorbonne, 75013 Paris}
\address[b]{Quarkslab SA, 13 rue Saint-Ambroise, 75011 Paris}
\address[c]{CEREMADE, CNRS, UMR 7534, Universit\'e Paris-Dauphine, PSL University, 75016 Paris}

\begin{abstract}
In this paper, we present a novel algorithm to address the \emph{Network Alignment problem}. It is inspired from a previous message passing framework of \citet{bayati_algorithms_2009} and includes several modifications designed to significantly speed up the message updates as well as to enforce their convergence. Experiments show that our proposed model outperforms other state-of-the-art solvers. Finally, we propose an application of our method in order to address the \emph{Binary Diffing problem}. We show that our solution provides better assignment than the reference differs in almost all submitted instances and outline the
importance of leveraging the graphical structure of binary programs.
\end{abstract}


\cortext[cor1]{Corresponding author.}
\end{frontmatter}

%\correspondingauthor[*]{Corresponding author. Tel.: +0-000-000-0000 ; fax: +0-000-000-0000.}
\email{elie.mengin@gmail.com}

\section{Introduction}
The problem of finding a relevant one-to-one correspondence between the nodes of two graphs may be known as the \emph{Network Alignment problem} (NAP). It has a wide variety of
applications such as image recognition \cite{zaslavskiy_path_2009,
  horaud_rigid_2011, zhou_factorized_2012}, ontology alignment
\cite{bayati_algorithms_2009}, social network analysis
\cite{zhang_final:_2016, kollias_network_2012} or protein interaction analysis
\cite{klau_new_2009, el-kebir_lagrangian_2011}. Multiple formal definitions have been proposed. For instance, one may only consider the node
content and aim at finding the mapping with maximum overall similarity
score. Such formulation reduces to the \emph{Maximum Weight Matching problem}
(MWM) \cite{bayati_maximum_2005}. On the contrary, one may only focus on the
graph topologies and search for the alignment with maximum induced overlapping
edges. In this case, the problem is known as the \emph{Maximum Common Edge Subgraph
  problem} (MCS) \cite{bahiense_maximum_2012}. In this paper, we propose a mixed formulation: given any two directed attributed graphs $A$ and $B$, and
two measures of similarity on both nodes and edges, find the one-to-one
mapping that maximizes the sum of similarity of both matched nodes and induced
edges. In other words, we are seeking the assignment that maximizes a
linear combination of MWM and MCS.

Several methods have been proposed to address this problem. Among them,
NetAlign \cite{bayati_algorithms_2009} introduces a complete message passing
framework based on \emph{max-product belief propagation}. In the present
paper, we propose several modifications to this algorithm in order to significantly speed up the computation and to control the messages
convergence. We show that these modifications provide better assignments than the original model as well as other state-of-the-art solvers in
most problem instances. Finally, we show that our method could be used to efficiently retrieve the differences between two binary executables.

The rest of this paper is organized as follows. Section \ref{sec:formulation}
introduces in more details the network alignment problem and reviews some
existing solutions. The original model of NetAlign as well as our proposed
optimizations are described in Section \ref{sec:framework}. Finally, Section
\ref{sec:evaluation} is dedicated to the experimental 
evaluation of our method. 

\section{Problem formulation} \label{sec:formulation} Let us consider any two
directed attributed graphs $A = (V_A, E_A)$ and $B = (V_B, E_B)$, where
$V_A = \{1, \dots, n\}$ are the vertices of $A$ (resp.
$V_B=\{1', \dots, m'\}$ for $B$) and
$E_A=\{(i, j) | i, j \in V_A^2, i \neq j\}$ are the edges of $A$
(resp. $E_B=\{(i', j') | i', j' \in V_B^2, i' \neq j'\}$ for $B$). Without
loss of generality, we assume that none of the graph includes self-loops (they
can be considered as node attributes if necessary).

We assume given two arbitrary non-negative measures of similarity on both
nodes and edges, $\sigma_V: i, i' \in V_A \times V_B \to \mathbb{R}_+$ and
$\sigma_E: (i, j), (i', j') \in E_A \times E_B \to \mathbb{R}_+$. Using these
measures, we may encode all pairwise node similarity scores into a flatten
vector $\mathbf{p} \in \mathbb{R}_+^{|V_A| \times |V_B|}$ such that $p_{ii'} =
\sigma_V(i, i')$, as well as the similarity of all potential induced edges
into matrix $Q \in  \mathbb{R}_+^{|V_A|^2 \times |V_B|^2}$ such that
$Q_{ii'jj'} = \sigma_E((i, j), (i', j'))$ if and only if $(i, j) \in E_A$ and
$(i', j') \in E_B$, and $0$ otherwise. Finally, we describe any one-to-one
node mapping through a binary vector $\mathbf{x} \in \{0,1\}^{|V_A| \times
  |V_B|}$ for which $x_{ii'}=1$ if and only if node $i$ in $A$ is matched with
node $i'$ in $B$. 

Given these definitions, the network alignment problem consists in solving the following constrained quadratic program:
\begin{equation}
	\tag{NAP}
	\label{def:NAP}
		\mathbf{x}^* =  \underset{\mathbf{x}}{\arg\max}\enspace
		\alpha \mathbf{x}^T \mathbf{p} + (1 - \alpha) \mathbf{x}^T Q
                \mathbf{x} \quad
		\text{ subject to } \enspace \forall i \in V_A, \sum_{j' \in V_B} x_{ij'}
                \le 1  \text{ and } \forall i' \in V_B, \sum_{j \in V_A} x_{ji'} \le 1,
\end{equation}
where $\alpha \in [0, 1]$ is an arbitrary constant that determines the trade-off between node and edge similarity.

This problem is a generalization of the \emph{Quadratic Assignment problem} and, as such, is known to be NP-complete and even APX-hard \cite{sahni_p-complete_1976}. Though exact algorithms exist, they rapidly become intractable as the number of vertices rises \cite{burkard_quadratic_1998}. In practice, the computation of the NAP for graphs of more than a hundred nodes must be approximated. In the rest of this section, we review some of the existing approaches proposed to address the problem.

\subsection*{Related work}
Amongst the first approaches to approximate the NAP are the spectral methods that can be distinguished into two main categories. On one hand, spectral matching approaches are based on the idea that similar graphs share a similar spectrum \cite{umeyama_eigendecomposition_1988}. Thus, they aim at best aligning the (leading) eigenvectors of the two affinity matrices (or Laplacians) \cite{horaud_rigid_2011, patro_global_2012}. On the other hand, PageRank methods approximates the NAP through an \emph{eigenvalue problem} over the matrix $Q$ \cite{singh_global_2008}. The idea consists in computing the principal eigenvectors of $Q$, and to use it as a similarity score of every possible correspondences. The resulting assignment can then be computed using conventional MWM solvers. Over the years, several improvements have been proposed to enhance the procedure \cite{kollias_network_2012, nassar_low_2018, feizi_spectral_2020, zhang_final:_2016}.

Other common approaches propose to directly address the quadratic program by means of relaxations. The most common convex relaxation consists in extending the solution set to the set of doubly stochastic matrices. The relaxed problem can then be exactly solved using convex-optimization solvers, and is finally projected into the set of permutation matrices to provide an integral assignment. However, when the solution of the convex program is far from the optimal permutation matrix, the final projection may result in an incorrect mapping \cite{lyzinski_graph_2016}. Other approaches make use of a concave \cite{zaslavskiy_path_2009} or indefinite \cite{vogelstein_fast_2014} relaxations. The induced programs are generally much harder to solve but yield better results when properly initialized. Note that most methods use a combination of both relaxations \cite{zhou_factorized_2012, zhang_kergm_2019}.

Several other methods are based on a linearization of the NAP objective function. The idea is to reformulate the quadratic program into an equivalent linear program, and to solve it using conventional (mixed-integer) linear programming solvers. However, this reformulation usually requires the introduction of many new variables and constraints, and computing the exact solutions of the linear program may become prohibitively expensive. In most cases, relaxations must also be introduced. A successful method, based on Adams and Johnson linearization and using a Lagrangian dual relaxation have been proposed by \citet{klau_new_2009} and later improved by \citet{el-kebir_lagrangian_2011}.

Finally, a message passing framework has shown promising results \cite{bayati_algorithms_2009}. It is directly derived from a previous model that provided important results on solving the MWM using \emph{max-product belief propagation} \cite{bayati_maximum_2005}. In our work, we chose to apply this model to our use case.

Note that an important limitation to the NAP is the size of the matrix $Q$ that grows quartically with the size of the graphs. This memory requirement may become prohibitive for relatively large graphs encountered in many real-world problems. It is mainly due to the intrinsic nature of the problem which requires that every potential correspondence is evaluated with regards to other candidates, in order to take into account its topological consistency. In practice, most methods are designed to efficiently exploit the potential sparseness of the matrix $Q$. Therefore, they generally apply to sparse graphs only. Moreover, several approaches propose to also restrain the number of potential candidates \cite{el-kebir_lagrangian_2011, bayati_algorithms_2009} and thus the problem complexity. This pre-selection may rely on prior knowledge or on arbitrary decision rules. It mostly aims at preventing the algorithm to compute the assignment score of highly improbable correspondences. The framework we introduce in the next section make use of this interesting feature.

\section{Network alignment via Max-Product Belief
  Propagation} \label{sec:framework}
In this section, we first recall the original model of \citet{bayati_algorithms_2009} and then introduce our proposed improvements. More details about the complete framework and practical implementation of NetAlign can be found in \cite{bayati_message-passing_2013}.

\subsection{Original model} \label{framework2}
The most common reformulation of a constrained optimization program into
probabilistic graphical models usually requires the introduction of a
factor-graph which assigns maximum probability to the solution of the
program. For more details about factor-graph graphical models, we refer the
reader to \citet{kschischang_factor_2006}. In order to design such graphical
model, we must encode both the objective function and the constraints of
\ref{def:NAP} into an equivalent probability distribution. This encoding is
done through the factorization of several functions (function nodes) over the
different variables of the program (variable nodes). In this form, the mode of
the distribution, can be efficiently computed (at least approximated) using
the \emph{max-product algorithm}. In the following, we introduce the
factor-graph designed to address \ref{def:NAP}. 

\subsubsection*{The factor-graph}
We first define the set of variable nodes $X =\{X_{ii'} \in \{0,1\}, ii' \in V_A \times V_B \}$. These variables record the correspondences belonging to the current mapping. We then introduce the different function nodes of our graphical model. We distinguish factors providing the energy to the objective function from factors encoding the program constraints.

On one hand, the objective function is encoded via two sets of function nodes $c_{ii'}: \{ 0, 1 \} \rightarrow \mathbb{R}^+$ and $c_{ii'jj'}: \{ 0, 1 \}^2 \rightarrow \mathbb{R}^+$, such that $\forall ii' \in V_A \times V_B, c_{ii'}(x_{ii'}) = e^{\alpha x_{ii'} p_{ii'}}$, and $\forall ii', jj' \in \left( V_A \times V_B \right)^2, c_{ii'jj'}(x_{ii'}, x_{jj'}) = e^{(1 - \alpha) x_{ii'} Q_{ii'jj'} x_{jj'}}$

On the other hand, the hard-constraints of \ref{def:NAP} are encoded using $\{0,1\}$ Dirac measures $f_i: \{ 0, 1 \}^{|\partial f_i|} \rightarrow \{ 0, 1 \} $, and similarly for $g_{i'}$, such that:
\begin{align*}
& \forall i \in V_A, f_i(x_{\partial f_i}) =
	\begin{cases}
		1, & \text{ if } \sum_{j' \in V_B} x_{ij'} \leq 1.\\
		0, & \text{ otherwise}.
	\end{cases}
& \forall i' \in V_B, g_{i'}(x_{\partial g_{i'}}) =
  	\begin{cases}
    	1, & \text{ if } \sum_{j \in V_A} x_{ji'} \leq 1.\\
    	0, & \text{ otherwise}.
  	\end{cases}
\end{align*}
where we denote $x_{\partial f_i} = \{x_{ij'} \in \textbf{x}, j' \in V_B\}$, and similarly, $x_{\partial g_{i'}} = \{x_{ji'} \in \textbf{x}, j \in V_A\}$.

By factorizing all the function nodes, we obtain the probability distribution of our factor-graph:
\begin{equation}
	p_X(\textbf{x}) = \frac{1}{Z} \left[ \prod_{i=1}^n f_i(x_{\partial f_i}) \prod_{j=1}^m g_j(x_{\partial g_j}) \right] e^{\alpha \textbf{x}^T \textbf{p} + (1 - \alpha) \textbf{x}^T Q \textbf{x}},
	\label{def:likelihood}
\end{equation}
where the normalization constant $Z$ denotes the partition function of the
model. It is clear that the support of our model distribution \eqref{def:likelihood} is equivalent to the set of feasible solutions in \ref{def:NAP}. Furthermore, the vector $\textbf{x}$ with maximum probability corresponds to the optimal solution of the \ref{def:NAP}.

\subsubsection*{The message passing framework}
The main interest of the factor-graph \eqref{def:likelihood} is the ability to efficiently compute an approximation of its mode using the \emph{max-product algorithm}. Following the message passing framework proposed by \citet{pearl_reverend_1982}, and denoting by $m^{(t)}$ the value of the message $m$ after $t$ iterations, we may apply the following updates \cite{bayati_algorithms_2009} :
\begin{align*}
&
\begin{aligned}  
\mu_{X_{ii'} \rightarrow f_i}^{(t+1)}(x_{ii'}) &= \lambda_{c_{ii'} \rightarrow X_{ii'}}^{(t)}(x_{ii'}) \lambda_{g_{i'} \rightarrow X_{ii'}}^{(t)}(x_{ii'}) \prod_{jj'} \lambda_{c_{ii'jj'} \rightarrow X_{ii'}}^{(t)}(x_{ii'}) \\
\mu_{X_{ii'} \rightarrow g_{i'}}^{(t+1)}(x_{ii'}) &= \lambda_{c_{ii'} \rightarrow X_{ii'}}^{(t)}(x_{ii'}) \lambda_{f_{i'} \rightarrow X_{ii'}}^{(t)}(x_{ii'}) \prod_{jj'} \lambda_{c_{ii'jj'} \rightarrow X_{ii'}}^{(t)}(x_{ii'}) \\
\mu_{X_{ii'} \rightarrow c_{ii'jj'}}^{(t+1)}(x_{ii'}, x_{jj'}) &= \lambda_{c_{ii'} \rightarrow X_{ii'}}^{(t)}(x_{ii'}) \lambda_{f_{i'} \rightarrow X_{ii'}}^{(t)}(x_{ii'}) \lambda_{g_{i'} \rightarrow X_{ii'}}^{(t)}(x_{ii'}) \prod_{kk' \neq jj'} \lambda_{c_{ii'kk'} \rightarrow X_{ii'}}^{(t)}(x_{ii'})
\end{aligned}\\
&
\begin{aligned}
\lambda_{f_{i} \rightarrow X_{ii'}}^{(t)}(x_{ii'}) &= \max_{\textbf{x}_{\partial f_i \backslash \{ ii' \}}} f_i(\textbf{x}_{\partial f_i}) \prod_{j' \neq i'} \mu_{X_{ij'} \rightarrow f_i}^{(t)} (x_{ij'}),
&& \lambda_{c_{ii'} \rightarrow X_{ii'}}^{(t)}(x_{ii'}) = c_{ii'}(x_{ii'})\\
\lambda_{g_{i'} \rightarrow X_{ii'}}^{(t)}(x_{ii'}) &= \max_{\textbf{x}_{\partial g_{i'} \backslash \{ ii' \}}} g_{i'}(\textbf{x}_{\partial g_{i'}}) \prod_{j \neq i} \mu_{X_{ji'} \rightarrow g_{i'}}^{(t)} (x_{ji'}) 
&& \lambda_{c_{ii'jj'} \rightarrow X_{ii'}}^{(t)}(x_{ii'}) = \max_{x_{jj'}} c_{ii'jj'}(x_{ii'}, x_{jj'}) \mu_{X_{jj'} \rightarrow c_{ii'jj'}}^{(t)}(x_{jj'})
\end{aligned}
\end{align*}
Since $x_{ii'}$ are binary valued, computing any message $\lambda_{a
  \rightarrow b}(x_{ii'})$ for $x_{ii'} = 0$ and $x_{ii'} = 1$ is
redundant. Therefore, we may halve the computation cost, by only considering
its log-ratio  $m_{a \rightarrow b} = \log{\frac{\lambda_{a \rightarrow
      b}(1)}{\lambda_{a \rightarrow b}(0)}}$. Following this notation, it can be shown \cite{bayati_algorithms_2009} that the messages from the variable nodes to the function nodes introduced above simplify to:
\begin{align*}
	m_{f_{i} \rightarrow X_{ii'}}^{(t)} &= - \left( \max_{k' \neq i'} m_{X_{ik'} \rightarrow f_{i}}^{(t)} \right)_+,
	&&  m_{c_{ii'} \rightarrow X_{ii'}}^{(t)} = \alpha p_{ii'}\\
	m_{g_{i'} \rightarrow X_{ii'}}^{(t)} &= - \left( \max_{k \neq i} m_{X_{ki'} \rightarrow g_{i'}}^{(t)} \right)_+,
	&& m_{c_{ii'jj'} \rightarrow X_{ii'}}^{(t)} =  \left((1 - \alpha) Q_{ii'jj'} + m_{X_{jj'} \rightarrow c_{ii'jj'}}^{(t)} \right)_+ - \left(m_{X_{jj'} \rightarrow c_{ii'jj'}}^{(t)} \right)_+
\end{align*}
where we use the notations: $x_+ = \max(0, x)$. Consequently, the message-passing framework reduces to the following updates:
\begin{align}
	\label{def:x-messages}
	m_{X_{ii' } \rightarrow f_i}^{(t+1)} &= m_{c_{ii'} \rightarrow X_{ii'}}^{(t)} + m_{g_{i'} \rightarrow X_{ii'}}^{(t)} + \sum_{jj'} m_{c_{ii'jj'} \rightarrow X_{ii'}}^{(t)} \\
	\label{def:y-messages}	
	m_{X_{ii' } \rightarrow g_{i'}}^{(t+1)} &= m_{c_{ii'} \rightarrow X_{ii'}}^{(t)} + m_{f_{i} \rightarrow X_{ii'}}^{(t)} + \sum_{jj'} m_{c_{ii'jj'} \rightarrow X_{ii'}}^{(t)} \\
	\label{def:z-messages}	
	m_{X_{ii' } \rightarrow c_{ii'jj'}}^{(t+1)} &= m_{c_{ii'} \rightarrow X_{ii'}}^{(t)} + m_{f_{i} \rightarrow X_{ii'}}^{(t)} + m_{g_{i'} \rightarrow X_{ii'}}^{(t)} + \sum_{kk' \neq jj'} m_{c_{ii'kk'} \rightarrow X_{ii'}}^{(t)}
\end{align}

\subsection{Proposed modifications}

\subsubsection*{Solution assignment}
After each message-passing iteration, the algorithm must compute the current best solution based on the updated messages. In their work, \citet{bayati_algorithms_2009} proposed several mechanisms, called ''rounding strategies''. Unfortunately, all of them require to address an instance of the MWM problem. Even worse, according to the authors, the best rounding strategy requires to solve exactly two MWM problems. Though this can be done in reasonable time, proceeding to this computational step after each iteration seriously slows down the algorithm.

To overcome this issue, we propose another simple assignment procedure based on the current estimated ''max-marginals''. In fact, following the notation introduced in Section \ref{framework2}, and referring to the computation rules of \citet{pearl_reverend_1982}, we may estimate the max-marginal distribution of each variable node such that:
\begin{equation*}
	\hat{p}_{X_{ii'}}^{(t)}(x_{ii'}) \propto \lambda_{c_{ii'} \rightarrow X_{ii'}}^{(t)}(x_{ii'}) \lambda_{f_i \rightarrow X_{ii'}}^{(t)}(x_{ii'})  \lambda_{g_{i'} \rightarrow X_{ii'}}^{(t)}(x_{ii'}) \prod_{jj'} \lambda_{c_{ii'jj'} \rightarrow X_{ii'}}^{(t)}(x_{ii'})
\end{equation*}
After $t$ iterations, we may deduce the current best assignment from the sign of each max-marginal log-ratios $\log \frac{\hat{p}_{X_{ii'}}^{(t)}(1)}{\hat{p}_{X_{ii'}}^{(t)}(0)}$:
\begin{equation*}
	\hat{X}_{ii'} = \argmax_{x \in \{ 0, 1 \}} \hat{p}_{X_{ii'}}^{(t)}(x_{ii'}) 
  	= \text{sign} \left( m_{c_{ii'} \rightarrow X_{ii'}}^{(t)} + m_{f_i \rightarrow X_{ii'}}^{(t)} + m_{g_{i'} \rightarrow X_{ii'}}^{(t)} + \sum_{jj'} m_{c_{ii'jj'} \rightarrow X_{ii'}}^{(t)} \right)
\end{equation*}
Note that this mechanism may result in a partial mapping. Therefore, after the last iteration, i.e. when the updates converge or reach the maximum number of iterations, we propose to enhance the resulting assignment with less confident matches by solving a MWM problem on the estimated max-marginal log-ratio of each unmatched node $\hat{X}_{ii'}$.

\subsubsection*{Auction based $\epsilon$-complementary slackness}
A well known problem of the \emph{Max-Product algorithm} when running on loopy graphical models is that it is not guaranteed to converge. In fact, it may fall into infinite loops and oscillate between few states \cite{murphy_loopy_1999}. Therefore, most implementations include a mechanism that enforces convergence \cite{braunstein_learning_2006, frey_clustering_2007}. In their work, \citet{bayati_algorithms_2009} propose a damping factor to mitigate the updates over iterations. Once this damping is sufficiently low, the message updates become insignificant, and the algorithm converges.

In our work, we propose another mechanism based on the concept of
$\epsilon$-complementary slackness \cite{bertsekas_auction_1992}. This
relaxation has been originally proposed for the \emph{Auction algorithm} to
address MWM instances that admit multiple optimal solutions. The idea is to
prevent the saturation of the complementary slackness with a small constant
$\epsilon$ margin. This scheme not only breaks ties and ensures the
convergence but also provably finds to the optimal solution for an $\epsilon$ small
enough \cite{bertsekas_auction_1992}. Furthermore, for larger
$\epsilon$ values, it shows to generally provide near-optimal assignments in
much less computation time. Though very similar in its MWM version ($\alpha=1$), our model is quite different from an \emph{Auction algorithm} in the general case. In order to adapt the idea of $\epsilon$-complementary slackness to our message-passing scheme, we propose the following modifications of updates \eqref{def:x-messages} and \eqref{def:y-messages}:
\begin{align*}
& m_{f_{i} \rightarrow X_{ii'}}^{(t)} = - \left( \max_{k' \neq i'} m_{X_{ik'} \rightarrow f_{i}}^{(t)} \right)_+ - 1_{m_{X_{ii'} \rightarrow f_{i}}^{(t)} \neq \underset{k'}{\text{max }} m_{X_{ik'} \rightarrow f_{i}}^{(t)}} \epsilon
& m_{g_{i'} \rightarrow X_{ii'}}^{(t)} = - \left( \max_{k \neq i} m_{X_{ki'} \rightarrow g_{i'}}^{(t)} \right)_+ - 1_{m_{X_{ii'} \rightarrow g_{i'}}^{(t)} \neq \underset{k}{\text{max }} m_{X_{ki'} \rightarrow g_{i'}}^{(t)}}\epsilon
\end{align*}
In our experiments, this mechanism shows to strongly favor the messages convergence and thus reduce the number of required running iterations. More importantly, this scheme tends to improve the overall final assignment score in many cases. 

However, this relaxation suffers from an important drawback: the value of the
introduced $\epsilon$ must be chosen carefully. If set too small, the
mechanism cannot fully play its part and the algorithm may reach a maximum
number of iterations before converging. On the contrary, if $\epsilon$ is too high, the
algorithm tends to converge too quickly to a poor local optimum. In their work, \citet{bertsekas_auction_1992} propose an iterative
method, called $\epsilon$-scaling, to properly setup the relaxation. The idea
consists in repeatedly decreasing $\epsilon$ after the messages converged,
until it reaches a small enough value, known to provide an optimal
solution. In our work, we suggest the opposite scheme. The model starts with a
rather small $\epsilon$ that helps to softly break local ties. Then, as the
algorithm iterates, we propose to rise the relaxation value each time the
messages have not improved the current objective function for few
iterations. As $\epsilon$ rises, the messages are more and more likely to
escape their local optimum and to fall into another better one. As soon as the
current assignment improves, $\epsilon$ is set back to its original value,
such that the new local solutions can be carefully explored.

\section{Evaluation} \label{sec:evaluation}

The proposed evaluation of our method, named QBinDiff, is twofold. We first
analyze its performances as a NAP solver. To do so, we submit the exact same
problems to several state of the art solvers, and compare the computed
alignment scores, without any consideration about the underlying purpose of
the problem instance. Then, we evaluate the relevance of our solution in order
to address the binary diffing problem. Therefore, we compare the resulting
mappings to \emph{ground truth} assignments and evaluate each matching method
with respect to accuracy metrics. In all our experiments, we ran our method
with default parameters: $\epsilon=0.5$, and within a maximum number of $1000$
iterations. 

\subsection{Benchmark experiments}
We compare our method to four state-of-the-art NAP solvers and their
associated benchmarks (see Table \ref{tab:benchmark-dataset}): PATH \cite{zaslavskiy_path_2009}, NetAlign
\cite{bayati_algorithms_2009}, Natalie 2.0 \cite{el-kebir_natalie_2015}, and
Final \cite{zhang_final:_2016}. All these solvers have been configured with their default parameters.
In order to analyze the ability of each solver to provide proper solutions
both in terms of node similarity and edge overlaps, we tested different values
of the trade-off parameter $\alpha$. Notice that some problems include a similarity score matrix with several zero entries. In some models (ours, NetAlign, Natalie), those entries are considered as unfeasable matches whereas they are legal correspondences in others. These models would thus optimize the problem on a subset of all possible one-to-one mappings.

\begin{table}
\centering
\caption{Description of our benchmark dataset.}
\label{tab:benchmark-dataset}
\begin{tabular}{lllrrrr}
\toprule
 Source & $A$ & $B$ & $|V_A|$ & $|V_B|$ & $|E_A|$ & $|E_B|$\\
\colrule
 \citet{zaslavskiy_path_2009} & 1EWK & 1U19 & 57 & 59 & 3192 & 2974 \\ 
 \citet{el-kebir_natalie_2015} & dmela & scere & 9459 & 5696 & 25635 & 31261 \\ 
 \citet{zhang_final:_2016} & flickr & lastfm & 15436 & 12974 & 32638 & 32298 \\ 
 \citet{bayati_algorithms_2009} & lcsh & wiki & 1919 & 2000 & 3130 & 7808 \\ 
 \citet{zhang_final:_2016} & offline & online & 1118 & 3906 & 3022 & 16328 \\ 
\botrule
\end{tabular}
\end{table}

Our results show that our approach outperforms or nearly ties the other
existing methods on every problems (see Table \ref{tab:benchmark-results}). It
appears to provide better results on sparse graphs, while it may compute
slightly suboptimal assignments on the densest one (1EWK-1U19). It also seems to be the best
fitted to perform diffing at different arbitrary setting of the trade-off
parameter $\alpha$, even in the degenerated MCS case ($\alpha=0$), unlike most
other evaluated methods.

In terms of computing time, as expected, QBinDiff takes much less time to
approximate the NAP than NetAlign. Regarding other solvers, both Natalie and
Final run within comparable time while Path tends to be very expensive, and
may become prohibitive for larger problem instances. Note that it was not able
to provide a solution to the Flickr-Lastm problem when $\alpha=0.25$ within 8
days, and was considered timed-out. Of course these timings depends on the
quality of the implementation and should only be considered with respect to
their order of magnitude. We used the implementations provided by the authors
of other methods. 

\subsection{Binary Diffing experiments}
In a second set of experiments, we propose to evaluate the relevance of our
model in order to address the \emph{Binary Diffing problem}. Given two binary
executables $A$ and $B$, this problem consists in retrieving the
correspondence between the functions of $A$ and those of $B$ that best
describes their semantic differences. This problem can be reduced
to a network alignment problem over the call graphs of $A$ and $B$. 

The results of the alignment should be compared to some ground truth. We
computed it as follows. We first downloaded the official repository of a
program, then compiled the different available versions using GCC v7.5 for
x86-64 target architecture and with -O3 optimization level. Once 
extracted, each binary was stripped to remove all symbols, then disassembled
using IDA Pro v7.2\footnote{https://www.hex-rays.com/products/ida}, and
finally exported into a readable file with the help of
BinExport\footnote{https://github.com/google/binexport}. During the problem
statement, only plain text functions determined during the disassembly process
are considered. For each program, assuming that this extraction protocol provided us with $n$
different versions, we propose to evaluate our method in diffing all the
$\frac{n(n-1)}{2}$ possible pairs of different binaries.

For all these diffing
instances, we finally had to extract the ground truth assignments. We proceed
in two steps. We first manually determine what we think to be the function
mapping that best describes the modifications between two successive program
versions. This is done considering the binary symbols, as well as the explicit
commit descriptions. Then, we deduce all the remaining pairwise diffing
assignments by extrapolating the mappings from version to versions. Formally,
if we encode the mapping between $A_1$ and $A_2$ into a boolean matrix
$M_{A_1 \to A_2}$ such that ${M_{A_1 \to A_2}}_{ii'} = 1$ if and only if
function $i$ in $A_1$ is paired with function $i'$ in $A_2$, then, our
extrapolating scheme simply consists in computing the diffing correspondence
between $A_k$ and $A_n$ as follows:
$M_{A_k \to A_n} = \prod_{i=k}^{n-1} M_{A_i \to A_{i+1}}$. We applied this
extraction protocol to three well known open source programs, namely Zlib
\footnote{https://github.com/madler/zlib}, Libsodium
\footnote{https://github.com/jedisct1/libsodium} and OpenSSL
\footnote{https://github.com/openssl/openssl} from which we collected
respectively 18, 33 and 17 different binary versions and therefore 153, 528
and 136 diffing instances. Statistics describing our evaluation dataset are
given in Table \ref{tab:dataset}.

\begin{table}[htbp]
\centering
\caption{Resulting objective scores of each solver on different benchmark problems. The last column records the average computing time in seconds.}
\label{tab:benchmark-results}
\begin{tabular}{llrrrrrr}
\toprule
 Problem & Matcher      &      $\alpha=0.0$ &     $\alpha=0.25$ &      $\alpha=0.5$ &     $\alpha=0.75$ &      $\alpha=0.9$ &       Time \\
\colrule
 \multirow{6}{*}{1EWK-1U19}
 & QBinDiff               & 2890.000 & 2183.366 & 1473.608 &  764.758 &  \textbf{339.628} &     22.532 \\
 & Final                  & 2874.000 & 2169.750 & 1465.500 &  761.250 &  338.700 &     17.944 \\
 & Natalie                & \textbf{2896.000} & 2172.496 & 1448.992 &  725.488 &  291.385 &    852.921 \\
 & NetAlign               & \textbf{2896.000} & \textbf{2185.750} & \textbf{1474.016} &  \textbf{765.023} &  338.700 &   1620.431 \\
 & Path                   & \textbf{2896.000} & 2169.750 & 1465.500 &  761.250 &  338.700 &      0.700 \\
\colrule
 \multirow{6}{*}{dmela-scere}
 & QBinDiff               &  \textbf{255.000} &  \textbf{347.937} &  \textbf{441.554} &  \textbf{543.832} &  \textbf{616.475} &     32.784 \\
 & Final                  &  112.000 &  250.953 &  389.906 &  528.859 &  612.230 &     45.898 \\
 & Natalie                &  174.000 &  163.109 &  142.558 &  126.837 &  119.156 &      8.677 \\
 & NetAlign               &  224.000 &  332.401 &  431.732 &  543.442 &  615.557 &    115.397 \\
 & Path                   &   47.000 &  230.809 &  376.118 &  522.177 &  609.812 &  17951.033 \\
\colrule
 \multirow{6}{*}{flickr-lastfm}
 & QBinDiff               & \textbf{6144.000} & \textbf{6955.018} & \textbf{7775.695} & \textbf{8594.405} & \textbf{9118.512} &    412.266 \\
 & Final                  & 1980.000 & 3890.405 & 5800.810 & 7711.215 & 8857.458 &    156.849 \\
 & Natalie                & 6012.000 & 5164.405 & 4282.425 & 3417.638 & 2896.056 &     94.092 \\
 & NetAlign               & 5830.000 & 6742.407 & 7567.495 & 8427.405 & 9025.683 & 434531.487 \\
 & Path                   &   10.000 &       NA & 4316.110 & 7403.845 &      8845.276 & 387114.102 \\
 \colrule
 \multirow{6}{*}{lcsh-wiki}
 & QBinDiff               &  \textbf{632.000} &  \textbf{614.867} &  \textbf{612.690} &  605.062 &  614.086 &     36.344 \\
 & Final                  &  574.000 &  585.217 &  596.435 &  \textbf{607.652} &  \textbf{614.383} &     20.302 \\
 & Natalie                &  584.000 &  496.465 &  419.760 &  337.640 &  287.927 &      2.932 \\
 & NetAlign               &  506.000 &  547.398 &  552.335 &  584.123 &  610.774 &     52.317 \\
 & Path                   &  238.000 &  385.085 &  462.935 &  539.480 &  594.963 &   4332.587 \\
\colrule
 \multirow{6}{*}{offline-online}
 & QBinDiff               & \textbf{2608.000} & \textbf{2138.206} & \textbf{1678.450} & \textbf{1103.162} &  \textbf{812.036} &     95.696 \\
 & Final                  &   40.000 &  222.352 &  404.704 &  587.055 &  696.466 &     22.685 \\
 & Natalie                &  198.000 &  315.160 &  327.387 &  392.081 &  446.390 &   1553.770 \\
 & NetAlign               & 1772.000 & 1507.908 & 1352.685 & 1002.279 &  756.600 &  27971.596 \\
 & Path                   &  244.000 &  789.614 &  725.422 &  690.011 &  742.723 &  13881.264 \\
\botrule
\end{tabular}
\end{table}

\begin{table}
\centering
\caption{Description of our binary diffing dataset. The last five
  columns respectively record the number of different binary versions, the number of resulting diffing instances, the
  average number of functions and function calls and the average ratio of conserved functions in our manually extracted ground truth.}
\label{tab:dataset}
\begin{tabular}{lrrrrr}
\toprule
 Program	& Versions & Problems & $\overline{|V]}$	& $\overline{|E]}$	& GT ratio\\
\colrule
 Zlib		& 18	& 153 & 153		& 235	& 0.96  \\
 Libsodium	& 33	& 528 & 589		& 701	& 0.79  \\
 OpenSSL	& 17	& 136 & 3473	& 18563	& 0.72 \\
\botrule
\end{tabular}
\end{table}

As a baseline, we chose to compare to the two most common diffing tools: BinDiff and Diaphora. BinDiff uses a matching algorithm close to VF2 \cite{cordella_subgraph_1998} originally introduced to approximate the MCS whereas Diaphora proposes a different greedy assignment strategy that first matches the most similar functions and then searches for potential correspondences in the remaining ones. This mapping mechanism is known to provide $\frac{1}{2}$-approximate solutions to the MWM problem \cite{duan_linear-time_2014}.
Note that, in order to compare with NAP solvers, and because, in general, binary diffing favors recall over precision, QBinDiff is designed to produce a complete assignment and does not include a mechanism to limit the mapping of very unlikely correspondences
during computation.

Our experiments show that QBinDiff generally outperforms other matching
approaches in both alignment score and recall (see Table \ref{tab:acc}). In fact, our method appears to perform clearly better at
diffing more different programs, whereas it provides comparable solutions on
similar binaries. This highlights that the
local greedy matching strategies of both BinDiff and Diaphora are able to provide good
solutions on simple cases but generalize poorly on more difficult problem
instances. This results should be view as promising in the perspective of
diffing much more different binaries.

We reproduced our experiments with different trade-off parameters $\alpha$, in order to estimate which setup should be used such that the optimal assignment meets the ground truth. Our computations suggest that a trade-off around 0.75 is a fair choice though a slightly higher value could provide satisfying assignments as well, especially while diffing OpenSSL programs.

\begin{table}
\centering
\caption{Average normalized resulting scores for each matching method. Computing time in seconds.} 
\label{tab:acc}
\begin{tabular}{lllrrrrr}
\toprule
 Project & Matcher   &   Similarity &   Squares &   Objective &   Precision &   Recall &   Time \\
\colrule
\multirow{7}{*}{Zlib}
 & QBinDiff  &        0.929 &     0.807 &  \textbf{ 0.888} &       0.955 &    \textbf{0.995} &       0.245 \\
 & BinDiff   &        0.921 &     0.765 &   0.869 &       0.943 &    0.975 &       1.239 \\
 & Diaphora  &        0.863 &     0.655 &   0.792 &       \textbf{0.978} &    0.940 &      10.494 \\
 \cmidrule{2-8}
 & Final     &        0.929 &     0.784 &   0.881 &       0.948 &    0.986 &      18.469 \\
 & Natalie   &        0.587 &     0.812 &   0.663 &       0.901 &    0.603 &       0.354 \\
 & NetAlign  &        0.929 &     0.807 &   \textbf{0.888} &       0.955 &    \textbf{0.995} &      21.935 \\
 & Path      &        0.930 &     0.789 &   0.883 &       0.953 &    0.992 &       0.589 \\
\colrule
\multirow{7}{*}{Libsodium}
 & QBinDiff  &        0.706 &     0.604 &   \textbf{0.677} &       0.722 &    \textbf{0.880} &      6.616 \\
 & BinDiff   &        0.662 &     0.544 &   0.628 &       0.752 &    0.869 &       1.278 \\
 & Diaphora  &        0.605 &     0.470 &   0.567 &       \textbf{0.783} &    0.831 &      35.327 \\
 \cmidrule{2-8}
 & Final     &        0.710 &     0.455 &   0.636 &       0.686 &    0.829 &      30.264 \\
 & Natalie   &        0.424 &     0.565 &   0.462 &       0.596 &    0.438 &      36.984 \\
 & NetAlign  &        0.703 &     0.597 &   0.673 &       0.722 &    0.879 &     283.130 \\
 & Path      &        0.700 &     0.519 &   0.648 &       0.696 &    0.836 &      61.784 \\
\colrule
\multirow{7}{*}{OpenSSL}
 & QBinDiff  &        0.720 &     0.595 &   \textbf{0.643} &       0.605 &    \textbf{0.783} &     213.252 \\
 & BinDiff   &        0.643 &     0.501 &   0.553 &       0.572 &    0.681 &       3.944 \\
 & Diaphora  &        0.543 &     0.332 &   0.408 &       0.577 &    0.565 &     228.666 \\
 \cmidrule{2-8}
 & Final     &        0.719 &     0.355 &   0.487 &       0.476 &    0.627 &     264.397 \\
 & Natalie   &        0.620 &     0.589 &   0.601 &       0.599 &    0.668 &     733.802 \\
 & NetAlign  &        0.682 &     0.587 &   0.623 &       \textbf{0.625} &    0.755 &   15193.922 \\
 & Path      &        0.722 &     0.521 &   0.596 &       0.556 &    0.715 &    7667.840 \\
\botrule
\end{tabular}
\end{table}

\subsection{Limitations}
Though our method improves the state-of-the-art, some difficulties remain.

A first limitation of our approach concerns the design of the network
alignment itself. Indeed, the determination of the trade-off parameter
$\alpha$ is subject to arbitrary considerations and may have an important
impact on the resulting solution. Moreover, this trade-off certainly depends
on the density of the graphs since densest graphs mechanically include more
potential edge overlaps. Finally, the proposed model is designed to compute
complete assignments and therefore does not include any mechanism to optimize
the precision. This later could be done in introducing a penalty term $\zeta$
to the node similarity scores such  $p_{ii'} = \sigma_V(i, i') -
\zeta$. Correspondences with negative similarity scores would thus belong to
the final mapping if they induce enough topological similarity. 

Another difficulty of our approach is the determination of the relaxation
parameter $\epsilon$. As previously mentioned, its setting controls on a
trade-off between the quality of the solution and the speed at which the
messages converges. While the proposed solution gives very satisfactory
results, other rising schemes could be used to adapt the
parameter to the current solution during computation. We leave this
investigations to future work. 

Finally, our formulation is limited by its memory consumption associated to
the use of a quartic memory matrix $Q$. Though the proposed model enables to significantly reduce the
problem size by limiting the solution set to the most probable
correspondences, this relaxation inevitably induces information loss,
especially for large graphs where the relaxation must rise consequently. In
practice, graphs of several thousands of nodes can be handled
efficiently. For larger instances, a solution could consists in first
partitioning the graphs into smaller consistent subgraphs, and then proceed
the matching  among them. However such partition is not trivial and might result in
important alignment errors.

\section{Conclusion}
In this paper, we introduced a new algorithm to address the network alignment
problem. It leverages a previous model and includes new mechanism to enforce
message convergence as well as speed-up the computation. Moreover, it proved
to provide better assignments than the original version in almost all alignment instances. 

Our evaluation showed that our approach outperforms state of the art
solvers. It also appears to be very well fitted to compute proper solutions
for different trade-offs between node similarity and edge overlaps. We finally
proposed an application of our method to address the binary diffing
problem. Our experiments showed that our algorithm provides better assignments
than other existing approaches for most diffing instances. This result
suggests that the formulation of the binary diffing problem as a network
alignment problem is the correct approach. 

\bibliography{bibliography}

\begin{thebibliography}{30}
\expandafter\ifx\csname natexlab\endcsname\relax\def\natexlab#1{#1}\fi
\providecommand{\url}[1]{\texttt{#1}}
\providecommand{\href}[2]{#2}
\providecommand{\path}[1]{#1}
\providecommand{\DOIprefix}{doi:}
\providecommand{\ArXivprefix}{arXiv:}
\providecommand{\URLprefix}{URL: }
\providecommand{\Pubmedprefix}{pmid:}
\providecommand{\doi}[1]{\href{http://dx.doi.org/#1}{\path{#1}}}
\providecommand{\Pubmed}[1]{\href{pmid:#1}{\path{#1}}}
\providecommand{\bibinfo}[2]{#2}
\ifx\xfnm\relax \def\xfnm[#1]{\unskip,\space#1}\fi
%Type = Article
\bibitem[{Bahiense et~al.(2012)Bahiense, Manić, Piva and
  De~Souza}]{bahiense_maximum_2012}
\bibinfo{author}{Bahiense, L.}, \bibinfo{author}{Manić, G.},
  \bibinfo{author}{Piva, B.}, \bibinfo{author}{De~Souza, C.C.},
  \bibinfo{year}{2012}.
\newblock \bibinfo{title}{The maximum common edge subgraph problem: {A}
  polyhedral investigation}.
\newblock \bibinfo{journal}{Discrete Applied Mathematics}
  \bibinfo{volume}{160}, \bibinfo{pages}{2523--2541}.
%Type = Inproceedings
\bibitem[{Bayati et~al.(2009)Bayati, Gerritsen, Gleich, Saberi and
  Wang}]{bayati_algorithms_2009}
\bibinfo{author}{Bayati, M.}, \bibinfo{author}{Gerritsen, M.},
  \bibinfo{author}{Gleich, D.F.}, \bibinfo{author}{Saberi, A.},
  \bibinfo{author}{Wang, Y.}, \bibinfo{year}{2009}.
\newblock \bibinfo{title}{Algorithms for {Large}, {Sparse} {Network}
  {Alignment} {Problems}}, in: \bibinfo{booktitle}{Proceedings of the 2009
  {Ninth} {IEEE} {International} {Conference} on {Data} {Mining}},
  \bibinfo{publisher}{IEEE Computer Society}, \bibinfo{address}{USA}. pp.
  \bibinfo{pages}{705--710}.
%Type = Article
\bibitem[{Bayati et~al.(2013)Bayati, Gleich, Saberi and
  Wang}]{bayati_message-passing_2013}
\bibinfo{author}{Bayati, M.}, \bibinfo{author}{Gleich, D.F.},
  \bibinfo{author}{Saberi, A.}, \bibinfo{author}{Wang, Y.},
  \bibinfo{year}{2013}.
\newblock \bibinfo{title}{Message-{Passing} {Algorithms} for {Sparse} {Network}
  {Alignment}}.
\newblock \bibinfo{journal}{ACM Transactions on Knowledge Discovery from Data
  (TKDD)} \bibinfo{volume}{7}, \bibinfo{pages}{3:1--3:31}.
%Type = Inproceedings
\bibitem[{Bayati et~al.(2005)Bayati, Shah and Sharma}]{bayati_maximum_2005}
\bibinfo{author}{Bayati, M.}, \bibinfo{author}{Shah, D.},
  \bibinfo{author}{Sharma, M.}, \bibinfo{year}{2005}.
\newblock \bibinfo{title}{Maximum weight matching via max-product belief
  propagation}, in: \bibinfo{booktitle}{Proceedings. {International}
  {Symposium} on {Information} {Theory}, 2005. {ISIT} 2005.}, pp.
  \bibinfo{pages}{1763--1767}.
%Type = Article
\bibitem[{Bertsekas(1992)}]{bertsekas_auction_1992}
\bibinfo{author}{Bertsekas, D.P.}, \bibinfo{year}{1992}.
\newblock \bibinfo{title}{Auction algorithms for network flow problems: {A}
  tutorial introduction}.
\newblock \bibinfo{journal}{Computational Optimization and Applications}
  \bibinfo{volume}{1}, \bibinfo{pages}{7--66}.
%Type = Article
\bibitem[{Braunstein and Zecchina(2006)}]{braunstein_learning_2006}
\bibinfo{author}{Braunstein, A.}, \bibinfo{author}{Zecchina, R.},
  \bibinfo{year}{2006}.
\newblock \bibinfo{title}{Learning by {Message} {Passing} in {Networks} of
  {Discrete} {Synapses}}.
\newblock \bibinfo{journal}{Physical Review Letters} \bibinfo{volume}{96},
  \bibinfo{pages}{030201}.
%Type = Incollection
\bibitem[{Burkard et~al.(1998)Burkard, Çela, Pardalos and
  Pitsoulis}]{burkard_quadratic_1998}
\bibinfo{author}{Burkard, R.E.}, \bibinfo{author}{Çela, E.},
  \bibinfo{author}{Pardalos, P.M.}, \bibinfo{author}{Pitsoulis, L.S.},
  \bibinfo{year}{1998}.
\newblock \bibinfo{title}{The {Quadratic} {Assignment} {Problem}}, in:
  \bibinfo{editor}{Du, D.Z.}, \bibinfo{editor}{Pardalos, P.M.} (Eds.),
  \bibinfo{booktitle}{Handbook of {Combinatorial} {Optimization}:
  {Volume1}–3}. \bibinfo{publisher}{Springer US}, \bibinfo{address}{Boston,
  MA}, pp. \bibinfo{pages}{1713--1809}.
%Type = Inproceedings
\bibitem[{Cordella et~al.(1998)Cordella, Foggia, Sansone and
  Vento}]{cordella_subgraph_1998}
\bibinfo{author}{Cordella, L.P.}, \bibinfo{author}{Foggia, P.},
  \bibinfo{author}{Sansone, C.}, \bibinfo{author}{Vento, M.},
  \bibinfo{year}{1998}.
\newblock \bibinfo{title}{Subgraph {Transformations} for the {Inexact}
  {Matching} of {Attributed} {Relational} {Graphs}}, in:
  \bibinfo{editor}{Jolion, J.M.}, \bibinfo{editor}{Kropatsch, W.G.} (Eds.),
  \bibinfo{booktitle}{Graph {Based} {Representations} in {Pattern}
  {Recognition}}, \bibinfo{publisher}{Springer}, \bibinfo{address}{Vienna}. pp.
  \bibinfo{pages}{43--52}.
%Type = Article
\bibitem[{Duan and Pettie(2014)}]{duan_linear-time_2014}
\bibinfo{author}{Duan, R.}, \bibinfo{author}{Pettie, S.}, \bibinfo{year}{2014}.
\newblock \bibinfo{title}{Linear-{Time} {Approximation} for {Maximum} {Weight}
  {Matching}}.
\newblock \bibinfo{journal}{Journal of the ACM} \bibinfo{volume}{61},
  \bibinfo{pages}{1:1--1:23}.
%Type = Inproceedings
\bibitem[{El-Kebir et~al.(2011)El-Kebir, Heringa and
  Klau}]{el-kebir_lagrangian_2011}
\bibinfo{author}{El-Kebir, M.}, \bibinfo{author}{Heringa, J.},
  \bibinfo{author}{Klau, G.W.}, \bibinfo{year}{2011}.
\newblock \bibinfo{title}{Lagrangian relaxation applied to sparse global
  network alignment}, in: \bibinfo{booktitle}{Proceedings of the 6th {IAPR}
  international conference on {Pattern} recognition in bioinformatics},
  \bibinfo{publisher}{Springer-Verlag}, \bibinfo{address}{Delft, The
  Netherlands}. pp. \bibinfo{pages}{225--236}.
%Type = Article
\bibitem[{El-Kebir et~al.(2015)El-Kebir, Heringa and
  Klau}]{el-kebir_natalie_2015}
\bibinfo{author}{El-Kebir, M.}, \bibinfo{author}{Heringa, J.},
  \bibinfo{author}{Klau, G.W.}, \bibinfo{year}{2015}.
\newblock \bibinfo{title}{Natalie 2.0: {Sparse} {Global} {Network} {Alignment}
  as a {Special} {Case} of {Quadratic} {Assignment}}.
\newblock \bibinfo{journal}{Algorithms} \bibinfo{volume}{8},
  \bibinfo{pages}{1035--1051}.
%Type = Article
\bibitem[{Feizi et~al.(2020)Feizi, Quon, Recamonde-Mendoza, Médard, Kellis and
  Jadbabaie}]{feizi_spectral_2020}
\bibinfo{author}{Feizi, S.}, \bibinfo{author}{Quon, G.},
  \bibinfo{author}{Recamonde-Mendoza, M.}, \bibinfo{author}{Médard, M.},
  \bibinfo{author}{Kellis, M.}, \bibinfo{author}{Jadbabaie, A.},
  \bibinfo{year}{2020}.
\newblock \bibinfo{title}{Spectral {Alignment} of {Graphs}}.
\newblock \bibinfo{journal}{IEEE Transactions on Network Science and
  Engineering} \bibinfo{volume}{7}, \bibinfo{pages}{1182--1197}.
%Type = Article
\bibitem[{Frey and Dueck(2007)}]{frey_clustering_2007}
\bibinfo{author}{Frey, B.J.}, \bibinfo{author}{Dueck, D.},
  \bibinfo{year}{2007}.
\newblock \bibinfo{title}{Clustering by {Passing} {Messages} {Between} {Data}
  {Points}}.
\newblock \bibinfo{journal}{Science} \bibinfo{volume}{315},
  \bibinfo{pages}{972--976}.
%Type = Article
\bibitem[{Horaud et~al.(2011)Horaud, Forbes, Yguel, Dewaele and
  Zhang}]{horaud_rigid_2011}
\bibinfo{author}{Horaud, R.}, \bibinfo{author}{Forbes, F.},
  \bibinfo{author}{Yguel, M.}, \bibinfo{author}{Dewaele, G.},
  \bibinfo{author}{Zhang, J.}, \bibinfo{year}{2011}.
\newblock \bibinfo{title}{Rigid and {Articulated} {Point} {Registration} with
  {Expectation} {Conditional} {Maximization}}.
\newblock \bibinfo{journal}{IEEE Transactions on Pattern Analysis and Machine
  Intelligence} \bibinfo{volume}{33}, \bibinfo{pages}{587--602}.
%Type = Article
\bibitem[{Klau(2009)}]{klau_new_2009}
\bibinfo{author}{Klau, G.W.}, \bibinfo{year}{2009}.
\newblock \bibinfo{title}{A new graph-based method for pairwise global network
  alignment}.
\newblock \bibinfo{journal}{BMC Bioinformatics} \bibinfo{volume}{10},
  \bibinfo{pages}{S59}.
%Type = Article
\bibitem[{Kollias et~al.(2012)Kollias, Mohammadi and
  Grama}]{kollias_network_2012}
\bibinfo{author}{Kollias, G.}, \bibinfo{author}{Mohammadi, S.},
  \bibinfo{author}{Grama, A.}, \bibinfo{year}{2012}.
\newblock \bibinfo{title}{Network {Similarity} {Decomposition} ({NSD}): {A}
  {Fast} and {Scalable} {Approach} to {Network} {Alignment}}.
\newblock \bibinfo{journal}{IEEE Transactions on Knowledge and Data
  Engineering} \bibinfo{volume}{24}, \bibinfo{pages}{2232--2243}.
%Type = Article
\bibitem[{Kschischang et~al.(2006)Kschischang, Frey and
  Loeliger}]{kschischang_factor_2006}
\bibinfo{author}{Kschischang, F.R.}, \bibinfo{author}{Frey, B.J.},
  \bibinfo{author}{Loeliger, H.A.}, \bibinfo{year}{2006}.
\newblock \bibinfo{title}{Factor graphs and the sum-product algorithm}.
\newblock \bibinfo{journal}{IEEE Transactions on Information Theory}
  \bibinfo{volume}{47}, \bibinfo{pages}{498--519}.
%Type = Article
\bibitem[{Lyzinski et~al.(2016)Lyzinski, Fishkind, Fiori, Vogelstein, Priebe
  and Sapiro}]{lyzinski_graph_2016}
\bibinfo{author}{Lyzinski, V.}, \bibinfo{author}{Fishkind, D.E.},
  \bibinfo{author}{Fiori, M.}, \bibinfo{author}{Vogelstein, J.T.},
  \bibinfo{author}{Priebe, C.E.}, \bibinfo{author}{Sapiro, G.},
  \bibinfo{year}{2016}.
\newblock \bibinfo{title}{Graph {Matching}: {Relax} at {Your} {Own} {Risk}}.
\newblock \bibinfo{journal}{IEEE Transactions on Pattern Analysis and Machine
  Intelligence} \bibinfo{volume}{38}, \bibinfo{pages}{60--73}.
%Type = Inproceedings
\bibitem[{Murphy et~al.(1999)Murphy, Weiss and Jordan}]{murphy_loopy_1999}
\bibinfo{author}{Murphy, K.P.}, \bibinfo{author}{Weiss, Y.},
  \bibinfo{author}{Jordan, M.I.}, \bibinfo{year}{1999}.
\newblock \bibinfo{title}{Loopy belief propagation for approximate inference:
  an empirical study}, in: \bibinfo{booktitle}{Proceedings of the {Fifteenth}
  conference on {Uncertainty} in artificial intelligence},
  \bibinfo{publisher}{Morgan Kaufmann Publishers Inc.},
  \bibinfo{address}{Stockholm, Sweden}. pp. \bibinfo{pages}{467--475}.
%Type = Inproceedings
\bibitem[{Nassar et~al.(2018)Nassar, Veldt, Mohammadi, Grama and
  Gleich}]{nassar_low_2018}
\bibinfo{author}{Nassar, H.}, \bibinfo{author}{Veldt, N.},
  \bibinfo{author}{Mohammadi, S.}, \bibinfo{author}{Grama, A.},
  \bibinfo{author}{Gleich, D.F.}, \bibinfo{year}{2018}.
\newblock \bibinfo{title}{Low {Rank} {Spectral} {Network} {Alignment}}, in:
  \bibinfo{booktitle}{Proceedings of the 2018 {World} {Wide} {Web}
  {Conference}}, \bibinfo{publisher}{International World Wide Web Conferences
  Steering Committee}, \bibinfo{address}{Lyon, France}. pp.
  \bibinfo{pages}{619--628}.
%Type = Article
\bibitem[{Patro and Kingsford(2012)}]{patro_global_2012}
\bibinfo{author}{Patro, R.}, \bibinfo{author}{Kingsford, C.},
  \bibinfo{year}{2012}.
\newblock \bibinfo{title}{Global network alignment using multiscale spectral
  signatures}.
\newblock \bibinfo{journal}{Bioinformatics} \bibinfo{volume}{28},
  \bibinfo{pages}{3105--3114}.
%Type = Inproceedings
\bibitem[{Pearl(1982)}]{pearl_reverend_1982}
\bibinfo{author}{Pearl, J.}, \bibinfo{year}{1982}.
\newblock \bibinfo{title}{Reverend bayes on inference engines: a distributed
  hierarchical approach}, in: \bibinfo{booktitle}{Proceedings of the {Second}
  {AAAI} {Conference} on {Artificial} {Intelligence}}, \bibinfo{publisher}{AAAI
  Press}, \bibinfo{address}{Pittsburgh, Pennsylvania}. pp.
  \bibinfo{pages}{133--136}.
%Type = Article
\bibitem[{Sahni and Gonzalez(1976)}]{sahni_p-complete_1976}
\bibinfo{author}{Sahni, S.}, \bibinfo{author}{Gonzalez, T.},
  \bibinfo{year}{1976}.
\newblock \bibinfo{title}{P-{Complete} {Approximation} {Problems}}.
\newblock \bibinfo{journal}{Journal of the ACM (JACM)} \bibinfo{volume}{23},
  \bibinfo{pages}{555--565}.
%Type = Article
\bibitem[{Singh et~al.(2008)Singh, Xu and Berger}]{singh_global_2008}
\bibinfo{author}{Singh, R.}, \bibinfo{author}{Xu, J.}, \bibinfo{author}{Berger,
  B.}, \bibinfo{year}{2008}.
\newblock \bibinfo{title}{Global alignment of multiple protein interaction
  networks with application to functional orthology detection}.
\newblock \bibinfo{journal}{Proceedings of the National Academy of Sciences}
  \bibinfo{volume}{105}, \bibinfo{pages}{12763--12768}.
%Type = Article
\bibitem[{Umeyama(1988)}]{umeyama_eigendecomposition_1988}
\bibinfo{author}{Umeyama, S.}, \bibinfo{year}{1988}.
\newblock \bibinfo{title}{An {Eigendecomposition} {Approach} to {Weighted}
  {Graph} {Matching} {Problems}}.
\newblock \bibinfo{journal}{IEEE Transactions on Pattern Analysis and Machine
  Intelligence} \bibinfo{volume}{10}, \bibinfo{pages}{695--703}.
%Type = Article
\bibitem[{Vogelstein et~al.(2014)Vogelstein, Conroy, Lyzinski, Podrazik,
  Kratzer, Harley, Fishkind, Vogelstein and Priebe}]{vogelstein_fast_2014}
\bibinfo{author}{Vogelstein, J.T.}, \bibinfo{author}{Conroy, J.M.},
  \bibinfo{author}{Lyzinski, V.}, \bibinfo{author}{Podrazik, L.J.},
  \bibinfo{author}{Kratzer, S.G.}, \bibinfo{author}{Harley, E.T.},
  \bibinfo{author}{Fishkind, D.E.}, \bibinfo{author}{Vogelstein, R.J.},
  \bibinfo{author}{Priebe, C.E.}, \bibinfo{year}{2014}.
\newblock \bibinfo{title}{Fast {Approximate} {Quadratic} {Programming} for
  {Large} ({Brain}) {Graph} {Matching}}.
\newblock \bibinfo{journal}{arXiv:1112.5507 [cs, math, q-bio]} .
%Type = Article
\bibitem[{Zaslavskiy et~al.(2009)Zaslavskiy, Bach and
  Vert}]{zaslavskiy_path_2009}
\bibinfo{author}{Zaslavskiy, M.}, \bibinfo{author}{Bach, F.},
  \bibinfo{author}{Vert, J.P.}, \bibinfo{year}{2009}.
\newblock \bibinfo{title}{A {Path} {Following} {Algorithm} for the {Graph}
  {Matching} {Problem}}.
\newblock \bibinfo{journal}{IEEE Transactions on Pattern Analysis and Machine
  Intelligence} \bibinfo{volume}{31}, \bibinfo{pages}{2227--2242}.
%Type = Inproceedings
\bibitem[{Zhang and Tong(2016)}]{zhang_final:_2016}
\bibinfo{author}{Zhang, S.}, \bibinfo{author}{Tong, H.}, \bibinfo{year}{2016}.
\newblock \bibinfo{title}{{FINAL}: {Fast} {Attributed} {Network} {Alignment}},
  in: \bibinfo{booktitle}{Proceedings of the 22nd {ACM} {SIGKDD}
  {International} {Conference} on {Knowledge} {Discovery} and {Data} {Mining}},
  \bibinfo{publisher}{Association for Computing Machinery},
  \bibinfo{address}{New York, NY, USA}. pp. \bibinfo{pages}{1345--1354}.
%Type = Incollection
\bibitem[{Zhang et~al.(2019)Zhang, Xiang, Wu, Xue and
  Nehorai}]{zhang_kergm_2019}
\bibinfo{author}{Zhang, Z.}, \bibinfo{author}{Xiang, Y.}, \bibinfo{author}{Wu,
  L.}, \bibinfo{author}{Xue, B.}, \bibinfo{author}{Nehorai, A.},
  \bibinfo{year}{2019}.
\newblock \bibinfo{title}{{KerGM}: {Kernelized} {Graph} {Matching}}, in:
  \bibinfo{editor}{Wallach, H.}, \bibinfo{editor}{Larochelle, H.},
  \bibinfo{editor}{Beygelzimer, A.}, \bibinfo{editor}{Alché-Buc, F.d.},
  \bibinfo{editor}{Fox, E.}, \bibinfo{editor}{Garnett, R.} (Eds.),
  \bibinfo{booktitle}{Advances in {Neural} {Information} {Processing} {Systems}
  32}. \bibinfo{publisher}{Curran Associates, Inc.}, pp.
  \bibinfo{pages}{3335--3346}.
%Type = Inproceedings
\bibitem[{Zhou(2012)}]{zhou_factorized_2012}
\bibinfo{author}{Zhou, F.}, \bibinfo{year}{2012}.
\newblock \bibinfo{title}{Factorized graph matching}, in:
  \bibinfo{booktitle}{Proceedings of the 2012 {IEEE} {Conference} on {Computer}
  {Vision} and {Pattern} {Recognition} ({CVPR})}, \bibinfo{publisher}{IEEE
  Computer Society}, \bibinfo{address}{USA}. pp. \bibinfo{pages}{127--134}.

\end{thebibliography}
\bibliographystyle{elsarticle-harv}

\end{document}